\def\year{2023}\relax
\documentclass[letterpaper]{article} % DO NOT CHANGE THIS
\usepackage{aaai24}  % DO NOT CHANGE THIS
\usepackage{times}  % DO NOT CHANGE THIS
\usepackage{helvet}  % DO NOT CHANGE THIS
\usepackage{courier}  % DO NOT CHANGE THIS
\usepackage[hyphens]{url}  % DO NOT CHANGE THIS
\usepackage{graphicx} % DO NOT CHANGE THIS
\def\UrlFont{\rm}  % DO NOT CHANGE THIS
\usepackage{natbib}  % DO NOT CHANGE THIS AND DO NOT ADD ANY OPTIONS TO IT
\usepackage{caption} % DO NOT CHANGE THIS AND DO NOT ADD ANY OPTIONS TO IT
\usepackage{times}  % DO NOT CHANGE THIS
\usepackage{helvet}  % DO NOT CHANGE THIS
\usepackage{courier}  % DO NOT CHANGE THIS
\usepackage[hyphens]{url}  % DO NOT CHANGE THIS
\usepackage{graphicx} % DO NOT CHANGE THIS
\def\UrlFont{\rm}  % DO NOT CHANGE THIS
\usepackage{natbib}  % DO NOT CHANGE THIS AND DO NOT ADD ANY OPTIONS TO IT
\usepackage{caption} % DO NOT CHANGE THIS AND DO NOT ADD ANY OPTIONS TO IT
\DeclareCaptionStyle{ruled}{labelfont=normalfont,labelsep=colon,strut=off} % DO NOT CHANGE THIS
\usepackage{bm}
\usepackage[ruled,vlined,linesnumbered]{algorithm2e} 
\usepackage{amsmath}
\usepackage{amssymb}
\DeclareMathOperator*{\E}{\mathbb{E}}

\usepackage{flexisym}
\usepackage{multirow}

\usepackage{multirow}
\usepackage{enumitem}
\usepackage{subcaption}

\newtheorem{problem}{Problem}
\newcommand{\commentout}[1]{%
}
\SetKwComment{Comment}{/* }{ */}

\usepackage{xcolor}
\usepackage{times}
\usepackage{soul}
\usepackage{url}
\usepackage[utf8]{inputenc}
\usepackage{graphicx}

\usepackage{booktabs}
\usepackage{amsmath}
\usepackage{enumitem}
\usepackage{bm}
\usepackage{subcaption}

\usepackage{algorithm2e}

\usepackage{xcolor}

\title{Cascade-based Randomization for \\ Inferring Causal Effects under Diffusion Interference
}

\author {
    Zahra Fatemi\textsuperscript{\rm 1}, 
    Jean Pouget-Abadie\textsuperscript{\rm 2}, 
    Elena Zheleva\textsuperscript{\rm 1}
}
\affiliations {
    \textsuperscript{\rm 1}University of Illinois Chicago\\
    \textsuperscript{\rm 2}Google Research\\
    zfatem2@uic.edu, jeanpa@google.com, ezheleva@uic.edu
}

\SetKwSty{textnormal}{}{}
\begin{document}

%\settopmatter{printacmref=false}
\maketitle


\begin{abstract}
  The presence of interference, where the outcome of an individual may depend on the treatment assignment and behavior of neighboring nodes, can lead to biased causal effect estimation. 
  Current approaches to network experiment design focus on limiting interference through cluster-based randomization, in which clusters are identified using graph clustering, and cluster randomization dictates the node assignment to treatment and control. 
  However, cluster-based randomization approaches perform poorly when interference propagates in cascades, whereby the response of individuals to treatment propagates to their multi-hop neighbors.
  When we have knowledge of the cascade seed nodes, we can leverage this interference structure to mitigate the resulting causal effect estimation bias.
 With this goal, we propose a cascade-based network experiment design that initiates treatment assignment from the cascade seed node and propagates the assignment to their multi-hop neighbors to limit interference during cascade growth and thereby reduce the overall causal effect estimation error. Our extensive experiments on real-world and synthetic datasets demonstrate that our proposed framework outperforms the existing state-of-the-art approaches in estimating causal effects in network data.

\end{abstract}

\section{Introduction}
Randomized Controlled Trials (RCTs), also known as A/B tests, are considered the gold standard for inferring causality \citep{antman-ama92,aral-ms11,bakshy-ec12,ho-slr17,betts-ajcn14,kazdin-pr22}. Through randomization, the experimenter can ensure that treatment and control assignments are independent of other variables. 
As a motivating example, consider a company that aims to assess the effectiveness of promoting its products through viral marketing (also known as word-of-mouth marketing) using an RCT. The campaign under consideration offers customers who have purchased a new phone (to which we refer as "influential nodes") a bill discount in exchange for sharing positive feedback about the product on their social media platforms.
To infer the causal effect of the campaign on product adoption, influential nodes are randomly split into a treatment group and a control group. The treatment group is exposed to the treatment (i.e., discount offer) and the control group is not. Then the product adoption rate in the social network of the treatment group nodes is compared to that in the social network of the control group nodes.

One of the challenges with designing a randomized controlled trial like this is that it is hard to isolate parts of a social network for treatment and control without any interactions. This is especially true when considering that word of mouth can form multi-hop spread over the network. The presence of interference, where the outcome of a node in the control group can be influenced by the treatment nodes and vice versa,
%the problem of treatment ”spilling over” from a treated node to a control node through connections, 
breaks a common causal inference assumption, known as the Stable Unit Treatment Value Assumption (SUTVA) \cite{rubin-ep74,imbens-book15,kohavi-eml17}. To address this problem, network experiment design focuses on ensuring reliable causal effect estimation in RCTs for potentially interacting units. 
Such network experiments have a wide range of applications across various disciplines, such as marketing \cite{aral-ms11,eckles-jci16,holtz-arxiv20}, and the medical \cite{halloran-ijb12,shakya-bmj17} and social sciences \cite{zhang-pmr16,glanz-ped17}.

\textit{Cluster-based randomization} or two-stage randomization
approaches are prominent network experiment designs for minimizing interference (also known as spillover or network effects). They find
densely connected clusters of nodes and assign treatment and control at the cluster level \cite{ugander-kdd13,eckles-jci17,saveski-kdd17,pouget-kdd18,ugander-arxiv20,fatemi-icwsm20}. Despite their advantage in mitigating interference, these approaches assume 1-hop spillover (i.e., interference between immediate neighbors). However, a piece of information (e.g., about a new product or a meme) can propagate from nodes to their multi-hop neighbors over time through a cascade of influence in the network. Cascades are natural phenomena in social networks and are the consequence of network interactions. In the running example, early adopters can trigger a large word-of-mouth cascade in the network by
influencing their friends on the social network to adopt the product, and their friends would influence their friends of friends, and so on.
The presence of a cascade can intensify interference in network experiments because each individual gets exposed to the treatment and outcomes of not only its immediate neighbors but also its multi-hop neighbors over time. 
Various aspects of network cascades have been studied in the research literature,  from cascade growth prediction \cite{cheng-www14,li-www17} to underlying network prediction \cite{netrapalli-met12,daneshmand-icml14,pouget-icml15}, and influence maximization \cite{chen-kdd09,wang-dmk12,chen-plo13,zhao-neu16}.  However, unbiased estimation of causal effects in cascade models remains an open problem.

\textbf{Present Work.} Motivated by applications in viral marketing and information diffusion, we propose a network experiment design to mitigate interference in cascade models. We specifically focus on the Independent Cascade Model \cite{kempe-kdd03}, a stochastic diffusion model where the adoption of a new product by a neighbor leads to neighboring nodes adopting the product with an independent probability. Under the assumption that cascade seed nodes (e.g., influential nodes) are known, we develop \textit{\textbf{Cas}cade-\textbf{B}ased \textbf{R}andomization (CasBR)}, a network experiment design which assigns treatment starting from the cascade seed nodes and propagates the assignment to their multi-hop neighbors. The intuition is that by assigning cascade seed nodes and their multi-hop neighbors to the same treatment group, we can limit interference between treatment and control groups.
Prior research has primarily utilized linear interference models and emphasized clustering as the optimal design for mitigating interference \citep{eckles-jci17, saveski-kdd17, brennan-neurips22}. In contrast, our paper challenges this approach by investigating a new interference model, the linear cascade model, and demonstrating that clustering may not be the most effective solution in the presence of a multi-hop cascade of influence.
To summarize, this paper makes the
following main contributions:
\begin{itemize}
    \item We formulate the problem of causal effect estimation in the context of the Independent Cascade Model, specifically in scenarios where we are aware of the cascade seed nodes (i.e., initially active nodes) before conducting a controlled experiment.
    \item We empirically show that existing cluster-based randomization approaches lead to highly biased causal effect estimates when cascades of influence occur in the network.
    \item We propose a cascade-based framework, CasBR, that leverages cascade seed nodes to create clusters of seeds and their multi-hop neighbors, with the aim of limiting interference in network experiments on cascade models.
\end{itemize}

The rest of the paper is structured as follows. In Section 2, we review the background of causal effect estimation in network experiments. In Section 3, we define our data model, causal estimand of interest, and the problem that we address in the paper. In Section 4, we present our cascade-based network experiment design framework. In Section 5, we present the experimental setup and results in real-world and synthetic datasets. In Section 6, we conclude and discuss directions for future work.

\section{Related Work}
While it is straightforward to randomly assign nodes to treatment and control groups in i.i.d data, causal effect estimation in interacting units is challenging due to information spillover from treated nodes to control nodes \cite{eckles-pnas16,aral-ms11,bailey-nber19,shakya-bmj17}.
A growing body of research has focused on
minimizing the effects of spillover in network experiments including multilevel designs where treatment is applied with different proportions across the population \cite{hudgens-jas08, tchetgen-smm12}, 
model-based approaches under specific models of interference \cite{basse-aas15,aronow-ims17,toulis-icml13,choi-jasa17}, methodologies that exclude neighboring nodes from the experiment \cite{fatemi-mlg20,fatemi-fbd23}, and experimental designs in two-sided platforms \cite{fradkin-ms21,johari-ms22}. 

Several methods rely on graph clustering to deal with network interference \cite{eckles-jci17,pouget-biometrika19,backstrom-www11,aronow-jci13,ugander-jci23,candogan-cbrp21}. 
Cluster-based randomization approaches partition the graph into clusters and allocate treatment cluster by cluster. 
Ugander et al. propose a  methodology using graph clustering to deal with interference and reduce the variance of the causal effect estimators \cite{ugander-kdd13}.
Saveski et al. develop a cluster-based model with balanced clusters to test for the presence of network interference in large-scale experimentation platforms \cite{saveski-kdd17}. 
Saint-Jacques et al. develop an ego-network randomization approach where a cluster is comprised of an ego and its immediate neighbors to measure the network effects and relax strong modeling assumptions existing in prior works \cite{saint-arxiv19}. 
Recent studies show the advantage of cluster-based randomization approach in causal effect estimation in bipartite graphs under interference \cite{brennan-neurips22,harshaw-arxiv21}

A new line of research shows that there is
a trade-off between interference and selection bias
in cluster-based randomization approaches based on the chosen number of clusters \cite{fatemi-icwsm20,fatemi-fbd23}. This work develops a network experiment design framework that combines node matching with weighted graph clustering to jointly minimize interference and selection bias. Since our work is closest to cluster-based randomization approaches, we use these methods as baselines in our experiments. 
Inspired by \citet{kempe-kdd03}, there are numerous studies on cascades from an influence maximization perspective \cite{chen-kdd09,morone-nature15,bharathi-ine07}. This line of work focuses more on influence maximization than the estimation of causal effects when influence propagates. On the topic of estimation, there is another line of work that seeks to estimate the graphs along which cascade models propagate ~\cite{gomez2012inferring}.
However, in our paper, we assume the graph is known and we seek to estimate the impact of cascades on network experiments.
 Recently, \citet{poiitis-tweb22} have looked into the value of choosing different seed node set sizes on aggression diffusion modeling and found that different seed sizes only affect the duration of the diffusion process but not the accuracy of user aggression scores.

\section{Problem Setup}
In this section, we formally define the data model, the causal estimand, and the problem we address in this paper.

\subsection{Data Model}
We consider a graph $G=\{\mathbf{V},\mathbf{E}\}$ with a set of $n$ nodes $\mathbf{V}$ and a set of edges $\mathbf{E}=\{ e_{ij}\}$ where $e_{ij}$ denotes that there is an edge between node $v_i\in \mathbf{V}$ and node $v_j\in \mathbf{V}$. Each node $v_i \in \mathbf{V}$ has an associated binary outcome variable $Y_i \in {0, 1}$ and a treatment assignment variable $T_i \in {0,1,2 }$, with $T_i=1$ indicating that the node is in treatment, $T_i=0$ indicating that the node is in control, and $T_i=2$ indicating that the node is excluded from the experiment.
If node $v_i$ is activated (e.g., adopted the product), then $Y_i=1.$
We define $\mathbf{Z} \in\{0, 1, 2\}^n$ as the treatment assignment vector of all nodes.
We further denote $N_i$ as the set of neighbors of node $v_i$.

\subsection{Independent Cascade Model}
The \textit{Independent Cascade Model (ICM)}  is a stochastic information diffusion model where information spreads through a cascade across the nodes of a graph \citep{kempe-kdd03}. In this model, each node can be in one of three states: 1) inactive, representing nodes that have not yet been activated by any of their neighboring active nodes, 2) active, representing nodes that were activated in the previous time step, and can activate inactive nodes in the next step, and 3) activated, representing nodes that activated other nodes in previous time steps and can no longer activate any other nodes.
In the ICM, a cascade begins with a set of active seed nodes denoted by $S_0$, and at each time step $t$, a set of newly active nodes $X^{new}_{t-1}$ is determined based on the previous time step. Each newly activated node can attempt to activate its inactive neighboring nodes with a certain propagation/spillover  probability $p(v_i, v_j)$. If a node has multiple active neighbors in $X^{new}_{t-1}$, their attempts to activate the node are sequenced in an arbitrary order. The diffusion process continues until no further activations are possible. This paper assumes that the initial set of active cascade seed nodes $S_0$ is known.

Our cascade model includes four distinct spillover probabilities: 1) $p_{t-t}$, which represents the probability of a treatment node activating another treatment node; 2) $p_{c-t}$, the probability of a control node activating a treatment node; 3) $p_{c-c}$, the probability of a control node activating another control node; and 4) $p_{t-c}$, the probability of a treatment node activating a control node.

We now describe the interference model in network experiments on cascade models. In the first time step of cascade propagation, cascade seed nodes are activated. In section \ref{sec:ceg}, we elaborate on two distinct methods for selecting the cascade seed nodes.
 After the treatment assignment mechanism is applied to the graph nodes, the following scenarios can occur based on the treatment assignment of an active node and its neighbor:
\begin{enumerate}
\item If the node and its neighbor are in treatment, the neighbor gets activated with probability $p_{t-t}$.
\item If the node and its neighbor are in control, the neighbor gets activated with probability $p_{c-c}$.
    \item If the node is in control and its neighbor is in treatment, the neighbor gets activated with probability $p_{c-t}$.
    \item If the node is in treatment and its neighbor is in control, the node gets activated with probability $p_{t-c}$.
\end{enumerate}
The cascade propagation proceeds until all the m-hop neighbors of the seed nodes have been explored.

\subsection{Types of Causal Effects Under Interference}
The causal estimand of interest in this paper is the Total Treatment Effects (TTE), defined as the difference between the outcomes of individuals in two alternative universes, one in which everyone receives treatment ($\mathbf{z_1}=\{1\}^n$) and another where no one does ($\mathbf{z_0}=\{0\}^n$) \cite{ugander-kdd13,fatemi-icwsm20}:
\begin{equation}
TTE=\frac{1}{n} \sum_{v_i \in V} (Y_i(\mathbf{Z}= \mathbf{z_1})-Y_i(\mathbf{Z}= \mathbf{z_0})).
\label{eq:tte}
\end{equation}

The fundamental problem for causal inference is that, for any individual, we can observe only the outcome under treatment or control group but not in both. Therefore, the effect $\hat{\tau}$ of a treatment on an outcome is typically calculated by averaging outcomes over treatment and control groups via difference-in-means: $\hat{\tau}=\overline{Y_{V_1}}-\overline{Y_{V_0}}$  where $Y_{V_1}$ and $Y_{V_0}$ denote the outcomes of treatment and control nodes, respectively ~\cite{halloran-epi95,stuart-ss10,fatemi-icwsm20}.

In real-world scenarios and in the presence of interference, the measured TTE is a combination of direct treatment effects and peer effects.
\textit{Direct Treatment Effects (DTE)} is defined as the difference between the average outcomes of treated and untreated individuals due to the treatment alone and is measured as:
\begin{equation}
DTE(V)=\E_{v_i\in V}[Y_i|T_i=1]-\E_{v_i\in V}[Y_i|T_i=0].
\end{equation}

Average \textit{peer effects (PE)} is the average influence of peers' behavior on the unit's response to the treatment and is estimated as:
\begin{align*}
PE(V)=&\E_{v_i\in V}[Y_i|T_i=\omega,N_i.\bm{\pi}] \\
-&\E_{v_i\in V}[Y_i|T_i=\omega,N_i=\emptyset],
\end{align*}
 where $N_i.\bm{\pi}$ denotes the vector of treatment assignments to node $v_i$’s neighbors $N_i$ 
 %vector of treatment assignment of $v_i$'s neighbors 
 and $\omega$ shows the treatment assignment of $v_i$. Peer effects are divided into two categories \cite{fatemi-icwsm20}: 
 \begin{itemize}
     \item \textit{Allowable peer effects (APE)}: APE is defined as the peer effects between nodes within the same treatment group (e.g., two peers received a voucher) and is measured as:
   \begin{align}
APE(V)=& \E_{v_i\in V}[Y_i|T_i=\omega,N_i^\omega.\bm{\pi}] \nonumber\\
-
&\E_{v_i\in V}[Y_i|T_i=\omega,N_i^\omega=\emptyset],
\end{align}
where $N_i^\omega$ denotes the set of neighbors of $v_i$ in the same treatment group as node $v_i$ with treatment assignment set $N_i^\omega.\bm{\pi}$.
     \item \textit{Unallowable peer effects (UPE)}: UPE is defined as the peer effects between neighbors with different treatment assignments (e.g., two friends one received a voucher and the other one did not) and is measured as:
     \begin{align}
UPE(V)=&\E_{v_i\in V}[Y_i|T_i=\omega,N_i^{\hat{\omega}}.\bm{\pi}] \nonumber\\
-
&\E_{v_i\in V}[Y_i|T_i=\omega,N_i^{\hat{\omega}}=\emptyset].
\label{eq:upe}
\end{align}

 \end{itemize}
where $N_i^{\hat{\omega}}$ ($\hat{\omega}\neq \omega$) denotes the set of neighbors in a different treatment group with treatment assignment $N_i^{\hat{\omega}}.\bm{\pi}$.

In the presence of allowable and unallowable peer effects in the network experiment, the measured TTE is a combination of DTE, APE, and UPE:
\begin{align}
TTE=&DTE(\mathbf{V})+APE(\mathbf{V_1}) \notag \\
-&APE(\mathbf{V_0})+UPE(\mathbf{V_1})-UPE(\mathbf{V_0}).
\end{align}
Where $\mathbf{V}_1$ and $\mathbf{V}_0$ represent sets of treatment and control nodes, respectively. Since APE are a natural consequence of network interactions, 
the main focus of this paper is to design a network experiment to minimize interference ($UPE(\mathbf{V_1})\approx 0$ and $UPE(\mathbf{V_0}) \approx 0$) between nodes in treatment and control groups in such a way that the measured TTE represents an unbiased estimation of the true underlying causal effects.

\subsection{Problem Definition}
In this paper, we are interested in setting up a network experiment design to limit interference between treatment and control groups in such a way that $TTE \approx DTE(\mathbf{V})+APE(\mathbf{V_1})-APE(\mathbf{V_0})$.
More formally:
\begin{problem}Network experiment design for the Independent Cascade Model.
\label{problem-def} 
 Given a graph $G = (\bm{V},\bm{E})$, 
  a set of attributes \textbf{V.X} associated with the graph nodes,
 a set of propagation probabilities $\bm{E.P}$ associated with the graph edges, and a set of active cascade seed nodes $S_0$, we would like to find 
a treatment assignment vector $\mathbf{Z}$ of a population with two different subsets of nodes, the treatment nodes $\bm{V}_1\in\bm{V}$, and the control nodes $\bm{V}_0\in\bm{V}$
%, and nodes excluded from the experiment $\bm{V}_2\in\bm{V}$,
such that: 

\begin{itemize}
\item[a.] $\mathbf{V}_0 \cap \mathbf{V}_1 = \emptyset$;
\item[b.] $|\mathbf{V}_0|+|\mathbf{V}_1|$ is maximized;
\item[c.] $P(v_i \in \mathbf{V_1}) \times P(v_i \in \mathbf{V_0}) >0; $
%\item[c.] \fixme{$P(T_i|X_i) >0$;}
\item[d.] Multi-hop interference ($UPE(\bm{V}_1)-UPE(\bm{V}_0)$) is minimized.
\end{itemize}
\end{problem}
The objective of the first component is to identify a set of treatment and control nodes that do not overlap with each other. The second component is designed to include as many nodes as possible from the entire node set $\bm{V}$, in order to obtain enough assignments to estimate the desired causal effects. The third component, referred to as the probabilistic/positivity assumption, requires that all units in the population under study have a positive chance of receiving any of the possible treatments. The fourth component aims to reduce multi-hop spillover in the experiment. It is worth noting that for certain methods of inference, having a positive overlap between treatment groups might not be adequate, and increasing the number of possible assignments across different treatment levels may improve the accuracy of the results.

\section{Cascade-based Network Experiment Design}
In this section, we present our network experiment design framework that mitigates unallowable peer effects (Eq. \ref{eq:upe}) in cascade models and provides a more accurate estimation of the causal effects (Eq. \ref{eq:tte}).

\begin{figure*}
\centering
  \includegraphics[width=\linewidth]{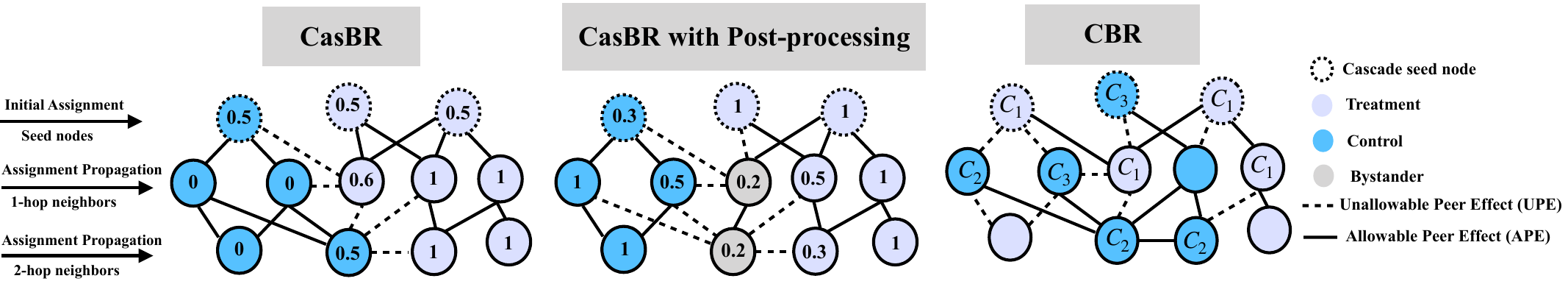}
\caption{The resulting node assignment of three different network experiment designs for minimizing interference in the Independent Cascade Model. 
The CBR approach leaves more edges between treatment and control nodes compared to our CasBR method (9 vs. 5 edges) which indicates more interference in the experiment.
}
\label{fig:icbr}
\end{figure*}
\subsection{Cascade-Based Randomization Framework}
With the goal of minimizing interference in the Independent Cascade Model, we propose \textit{\textbf{Cas}cade-\textbf{B}ased \textbf{R}andomization (CasBR)}, a network experiment design framework that uses the cascade seed nodes as the seeds of the treatment randomization procedure to reduce the interference during cascade propagation in network experiments. Given the initial seeds of a cascade, CasBR consists of two main steps:
%\vspace{-22}
\begin{enumerate}
    \item \textbf{Initial randomization}: In this step, the set of cascade seed nodes $\mathbf{S_0}$ are assigned to the treatment or control groups at random. This randomization allows the framework to create $|\mathbf{S_0}|$ clusters of treatment and control nodes with cascade seeds as the seed of each cluster. 
    \item \textbf{Assignment propagation}: In this step,  treatment assignment is propagated in a multi-hop manner in the network, to satisfy requirement (d) of Problem 1. First, the immediate neighbors of the seed nodes are assigned to treatment or control. 
     Considering the fraction of treated and untreated neighbors, each unassigned node at distance $m$ from a cascade seed node is either assigned to treatment with probability $p_t(v_i)$ measured as:
    \begin{equation}
    \label{eq:pt}
        p_t(v_i)=\frac{|treated_{ngb}(v_i)|}{|treated_{ngb}(v_i)|+|control_{ngb}(v_i)|},
    \end{equation}
     or assigned to control with probability $p_c(v_i) = 1-p_t(v_i)$. 
    $treated_{ngb}(v_i)$ stands for the set of treated neighbors of $v_i$ and $control_{ngb}(v_i)$ denotes the set of neighbors in control. 
    In order to avoid allocating all the budget to one treatment group, we create two distinct sets of unassigned neighbors of treatment and control nodes and then alternate between two sets. This fulfills requirement (a) of Problem 1.

    Once the immediate neighbors of the seed nodes have been examined, we proceed to repeat the randomization process on the unassigned 2-hop neighbors of the seed nodes. This process is then repeated until all the 1-hop, 2-hop, and m-hop neighbors of the seed nodes have been assigned to either the treatment or control group. Singletons (i.e., isolated nodes without any neighbors) are assigned randomly to either the treatment or control group. 
    This satisfies requirement (b) of Problem 1. Since each seed node is randomly assigned to the treatment or control group, and any node, whether seed or non-seed, could have been assigned to either treatment group, our approach fulfills requirements (c) of Problem 1.
   
\end{enumerate} 

\begin{algorithm}[t]
\SetKwInOut{Input}{Input}
\SetKwInOut{Output}{Output}
\Input{$G(\mathbf{V},\mathbf{E}),\mathbf{S_0}$}
\Output{$\mathbf{Z}=\{0,1\}^n$}
\DontPrintSemicolon
\caption{\mbox{Cascade-based Randomization CasBR(G)}}
\label{alg:icbr}
\BlankLine
$\text{Initialize } \mathbf{Z} \text{ by randomizing over cascade seeds } \mathbf{S_0};$\;
\While{$\{\exists \ v_i \in \mathbf{V}:$ $Z_j = \emptyset\}$}{
    $\mathbf{N_t} = \{ v_i \mid \exists \ e_{ji} \in \mathbf{E} \ \& \ Z_j=1 \ \& \ Z_i=\emptyset \}$\;
    $\mathbf{N_c} = \{ v_i \mid \exists \ e_{ji} \in \mathbf{E} \ \& \ Z_j=0 \ \& \ Z_i=\emptyset \}$\;
    \SetKw{KwShuffle}{Shuffle}
    \KwShuffle($\mathbf{N_t}$, $\mathbf{N_c}$)\;
    \While{$\{\mathbf{N_t} \neq \emptyset$ or $\mathbf{N_c} \neq \emptyset\}$}{
        \SetKw{KwPop}{Pop the first unassigned node}
        \KwPop $v_i$ from $\mathbf{N_t}$\;
        \SetKw{KwSetTreated}{Set of immediate neighbors of}
        $treated_{ngb} \gets$ \KwSetTreated $v_i$ assigned to treatment\;
        $control_{ngb} \gets$ \KwSetTreated $v_i$ assigned to control\;
        Calculate $p_t$ based on $treated_{ngb}$ and $control_{ngb}$ using Eq.\ref{eq:pt}\;
        Assign $v_i$ to treatment with probability $p_t$\;
        $Z_i \leftarrow 1$ if $v_i$ is assigned to treatment and $Z_i \leftarrow 0$ otherwise\;
        \KwPop $v_i$ from $\mathbf{N_c}$\;
        Repeat Lines 8-12\;
    }
}
\Return{$\mathbf{Z}$}
\end{algorithm}

\begin{algorithm}[h]
\SetKwInOut{Input}{Input}
\SetKwInOut{Output}{Output}
\Input{$G(\mathbf{V},\mathbf{E}), \alpha$}
\Output{$\mathbf{Z}=\{0,1,2\}^n$}
\DontPrintSemicolon
\caption{Post-processing}
\label{alg:icbrpost}
\BlankLine
$\mathbf{Z} = \text{CasBR}(G)$\;
\For{$v_i \in \mathbf{V}$}{
    \If{$\frac{|treated_{ngb}(v_i)|-|control_{ngb}(v_i)|}{{degree}(v_i)} < \alpha$}{
        $Z_i \leftarrow 2$\;
    }
}
\Return{$\mathbf{Z}$}
\end{algorithm}

    \textbf{Post-processing}:
The objective of the CasBR framework is to allocate nodes and their multi-hop neighbors to a similar treatment group. However, in some cases, nodes may have multiple neighbors from both treatment and control groups, referred to as \textit{bystander} nodes (e.g., gray nodes in Fig. \ref{fig:icbr}). By excluding these nodes from the experiment, we may be able to reduce interference between treatment and control. However, peer effects from bystander nodes to treatment and control nodes may still occur via their peers. To mitigate this issue, we can choose bystander nodes with a more similar distribution of treatment and control neighbors. By doing so, the peer effects of bystander nodes on the treatment and control groups can cancel each other out in the total treatment effect estimation, leading to a reduction in the bias of the measured causal effects. To identify bystander nodes, we compute the difference between the fraction of treatment and control neighbors of each node $v_i$, denoted by $\beta_i$. If $\beta_i$ is less than a threshold $\alpha$, the node is added to the set of bystander nodes. Lower values of $\alpha$ indicate higher similarity between the distribution of treatment and control nodes in the neighborhood of bystander nodes. The measurement of $\beta_i$ for each node $v_i$ is defined as:
\begin{equation}
\label{eq:beta}
\beta_i= \frac{|treated_{ngb}(v_i)|-|control_{ngb}(v_i)|}{degree(v_i)} <\alpha.
\end{equation}  

Algorithm \ref{alg:icbr} and \ref{alg:icbrpost} depict the CasBR and post-processing algorithms as pseudo-code. The CasBR algorithm assigns nodes in $\mathbf{V}$ to either treatment or control. In Line 1 of Algorithm \ref{alg:icbr}, the initial cascade seed nodes are randomly assigned to either treatment or control. In Lines 3-5, two sets of 1-hop (immediate) neighbors of treatment and control nodes are created and shuffled. In Lines 7-14, the probability of assigning each neighbor of seed nodes to treatment or control is calculated, and based on that, the node is assigned to treatment or control. The same process is then repeated for inactive 2-hop, ..., and m-hop neighbors of the cascade seeds to ensure that all nodes in the network are assigned to either treatment or control. In the worst-case scenario, m is equal to the diameter of the network which is the shortest distance between the two most distant nodes in the network.
During the post-processing step (Algorithm \ref{alg:icbrpost}), which is optional, a node is added to set $\mathbf{V}_2$ containing bystander nodes if the condition in Line 3 is met. 

Fig. \ref{fig:icbr} presents a visualization of the node assignment for three different network experiment designs, using a toy example with three cascade seed nodes, five 1-hop neighbors, and four 2-hop neighbors.
In the first step of our CasBR approach, we randomly assign the cascade seed nodes (shown as dashed circles) to treatment or control with a probability $p_t=0.5$. In the next step, we assign immediate unassigned neighbors of the seed nodes to either treatment or control with probability $p_t$ calculated with Eq. \ref{eq:pt}. The numbers in each circle of the CasBR graph represent that probability of assigning each node to treatment at different distances from the cascade seed nodes. Similar to the 1-hop neighbors, we conduct randomization on the 2-hop neighbors of the seed nodes. Dashed lines show the presence of unallowable peer effects between endpoint nodes.
The second experiment design is the same as the first one, except that we apply post-processing on CasBR and remove bystander nodes (shown as gray circles) from the experiment. The numbers in each circle of the post-processing graph are calculated using Eq. \ref{eq:beta} where we set $\alpha=0.3$.
In the CBR approach, we partition the toy graph into three clusters using the reLDG clustering algorithm \cite{nishimura-kdd13}. Each circle in the CBR graph is annotated with the assigned cluster number. Nodes that remain unassigned by the reLDG algorithm are represented by circles without any annotation. These nodes are randomly assigned to the treatment or control group.
The reason for poor clustering may be related to the fact that the algorithm is trying to find balanced clusters. This can lead to a situation where the algorithm prioritizes balancing the sizes of the clusters over finding densely connected clusters.
As shown in Fig. \ref{fig:icbr}, the CBR approach leads to a $57\%$ increase (9 vs. 5 dashed edges) in the number of edges between treatment and control nodes compared to the CasBR approach which implies a higher degree of interference in the experiment.

As our algorithm involves iterating over nodes and their neighbors to identify unassigned nodes, its complexity closely resembles that of a Breadth-First Search (BFS) algorithm. For a graph with $|\mathbf{V}|$ nodes and $|\mathbf{E}|$ edges,
the computational complexity of our algorithm is $O(|\mathbf{V}| +  |\mathbf{E}|)$.
In the worst-case scenario, the graph is a clique, and the complexity of the algorithm is dominated by the edge size which is $O(|\mathbf{V}|^2)$.

\section{Experiments}
In this section, we evaluate the performance of different methods in estimating causal effects. We first describe datasets used in our experiments and then discuss our baselines and results. 
\subsection{Semi-Synthetic Data Generation}
Since we do not have the underlying ground truth of the causal effects in real-world datasets, we rely on synthetic and semi-synthetic data generation for the evaluation of different network experiment designs.
\subsubsection{Dataset Description}
In our experiments, we consider 10 real-world datasets some of which are attributed. Table \ref{tb:datasets} represents the datasets' characteristics.  For weakly-connected networks, we compute the maximal shortest path as the network's diameter.
The \emph{Pubmed} datasets is a citation network with TF/IDF weighted word vectors attributes \cite{sen-aimag08}. The \emph{Hateful Users} dataset is a subset of Twitter's retweet graph, consisting of users annotated as hateful—i.e., users who post hateful and offensive content on social media—or not, as described in Ribeiro et al.~\cite{ribeiro-icwsm18}.  
\emph{Facebook} ego-network contains social circles on Facebook collected from survey participants using a Facebook app \cite{leskovec-neurips12}. \emph{Hamsterster} dataset includes the online social network of a group of
hamsters \cite{zheleva-snakdd08}. \emph{ArXiv} dataset contains the collaborations between authors who submitted their papers to the General Relativity and Quantum Cosmology category of e-print arXiv \cite{leskovec-kdd07-graph}. \emph{ACM} dataset is a paper authorship network from the ACM dataset with bag-of-words attributes for each paper.
\emph{Wiki} dataset is a word co-occurrence network constructed from the entire set of English Wikipedia pages \cite{cucerzan-emnlp07}. 
\emph{AMAP} is a product co-purchased network extracted from Amazon with product reviews encoded by bag-of-words as the attributes \cite{shchur-neurips18}.

\begin{figure*}[ht]
\centering
  \includegraphics[width=\linewidth]{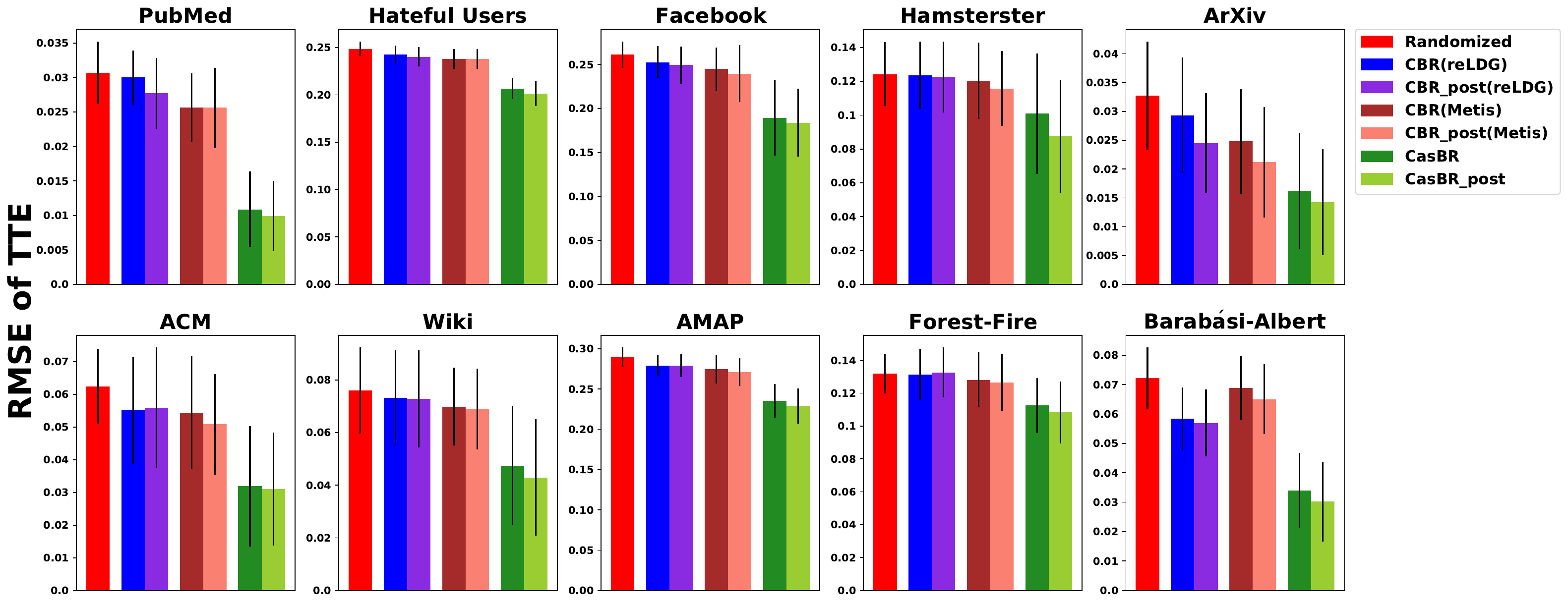}
\caption{ RMSE of total treatment effect estimation of different methods on real-world and synthetic datasets in the last time step of the cascade propagation; Bars show the standard deviation of the estimated effects. Seed nodes are selected using the random sampling method.
In all datasets, $10\%$ of the nodes are considered as the cascade seeds. CasBR and CasBR-post achieve the best performance in all datasets. 
}
\label{fig:interference}
\end{figure*}

\begin{table}[t]

\centering
\setlength\tabcolsep{3pt}
\begin{tabular}{lcccc}  
\toprule
 Dataset & Nodes & Edges & Attributes& Diameter\\
\midrule
 %Catster &73204  & 221293 \\
 Pubmed &19,717&44,325&500&17\\
 Hateful Users& 20,810&325,007& 523 & 11 \\
Facebook&4,039&88,234&-&8\\
 Hamsterster & 2,059 &10,943&6 &10\\
  ArXiv &5,242  &14,496&- &17\\
   ACM &3,025 &13,128& 1,870&20\\
 Wiki&2,405&17,981&4,973&9\\
AMAP&7,650&119,081&745&11\\
 
\bottomrule
\end{tabular}
\caption{Number of nodes, edges, attributes, and diameter of the real-world networks}
\label{tb:datasets}
\end{table}

We also generate two networks with 5,000 nodes using two different synthetic graph generators.
In the \textit{Barab$\acute{a}$si-Albert} model which generates random scale-free networks with preferential attachment, we set the parameter that controls the number of nodes a new node can attach to 3. In the \textit{Fire-Forest} model, a new node $v_i$ attaches to an existing node $v_j$ and then begins burning links outward from $v_j$, linking with forward ($p_f$) and backward ($p_b$) probabilities to any new node it discovers \cite{leskovec-kdd07-graph}. 
In our experiments, we set $p_f=0.35$ and $p_b=0.4$.

\subsubsection{Cascade Seed Selection}
\label{sec:ceg}
In cascade propagation, the seed selection process plays a crucial role, as the configuration of the propagation model, and the size of the cascade are intricately linked to the initial seed nodes' choice. 
%In the real-world scenario, it is hard to know the active cascade seed nodes.
We consider two methods to choose the cascade seed nodes: 1) \textit{random sampling} where a random subset of nodes are selected independently \cite{poiitis-tweb22}, and 2) \textit{NewGreedyIC} algorithm, a greedy influence maximization approach to find the smallest set of cascade seed nodes that could maximize the spread of influence in the network \cite{chen-kdd09}. The NewGreedyIC algorithm is a plausible way that viral marketing campaigns might use to choose their seed nodes with the goal of maximizing their product adoption. In the NewGreedyIC algorithm, we run 100 simulations and set the propagation probability to 0.01.

\subsubsection{True Total Treatment Effect Estimation}
To accurately measure the true underlying TTE, we consider two parallel universes, one where all nodes are receiving treatment and the other where all nodes are in the control group. Initially, we activate the cascade seed nodes and allow their activations to spread through the network. At each step, a newly activated node can activate its neighbors with probability $p_{t-t}$ and $p_{c-c}$ in the treatment and control universes, respectively. This process continues until all m-hop neighbors of each seed node are examined. We then calculate the difference between the average outcomes of nodes in the two universes over Q simulations, which indicates the true underlying TTE.

To quantify the strength of interference, we measure the difference between the average outcomes of treatment and control nodes in each time step of the cascade over Q simulations as estimated TTE in that time step. We report \textit{Root Mean Square Error} of total treatment effects measured as:
\begin{equation}
   RMSE= \sqrt{\frac{1}{Q}\sum_{q=1}^{Q} (\hat{\tau}_q-\tau_q)^2}
\end{equation}
where $\tau_q$ and $\hat{\tau}_q$ are the true and estimated TTE in simulation $q$, respectively. In our cascade generation, 
we set $Q=100$, $p_{t-t}=p_{t-c}=0.05$, and $p_{c-c}=p_{c-t}=0.02$. The experiments were conducted on an Ubuntu 20.04.4 server equipped with 128 CPUs, each running at 2000 MHz, and 256 GB of RAM.

\begin{table}[h]
\centering
%\small\addtolength{\tabcolsep}{-4pt}
\begin{tabular}{lcccc}  
\toprule
 Dataset & CasBR-post & CBR-post(reLDG)\\
\midrule
 %Catster &73204  & 221293 \\
 Pubmed&  11.6 &23.1 \\
 Hateful Users& 6.3&8.5 \\
Facebook& 17&28.2\\
 Hamsterster  &7.1&8.2\\
  ArXiv & 4.3&10.4\\
   ACM & 2.3 & 6.2\\
 Wiki&7.5 &11.6\\
AMAP& 5.9&7.8\\
Barab$\acute{a}$si-Albert&10.5&12.9\\
Forest-Fire&7.6&10.8\\
 
\bottomrule
\end{tabular}
\caption{Percentage of nodes excluded from the experiments using CasBR-post and CBR-post(reLDG) models}
\label{tb:post_pro}
\end{table}

\begin{figure*}[ht]
\centering
  \includegraphics[width=\linewidth]{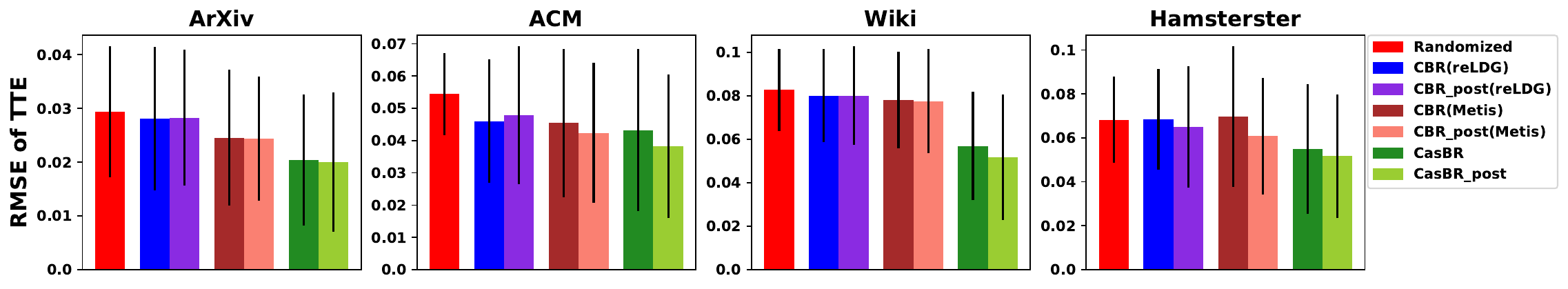}
\caption{ RMSE of total treatment effect estimation of different methods in the last time step of the cascade propagation; Bars show the standard deviation of the estimated effects. Seed nodes are selected using the random sampling method.
In all datasets, $10\%$ of the nodes are considered as the cascade seeds. We set $p_{t-t}=p_{t-c}=0.07$, and $p_{c-c}=p_{c-t}=0.05$.
}
\label{fig:interference_prob}
\end{figure*}

\begin{figure*}[h]
\begin{subfigure}{.49\textwidth}
  \centering
  % include second image
  \includegraphics[width=\textwidth]{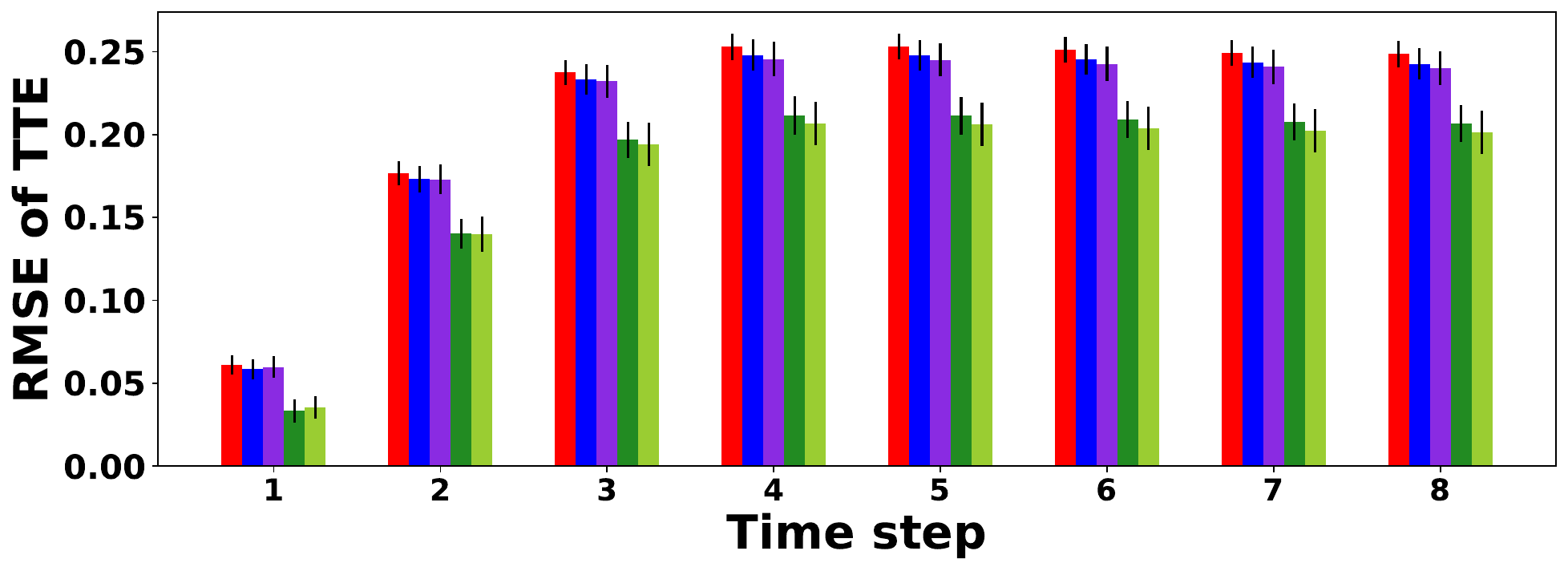}  
  \caption{Hateful Users}
  \label{fig:hateful}
\end{subfigure}
\begin{subfigure}{.49\textwidth}
  \centering
  % include second image
  \includegraphics[width=\textwidth]{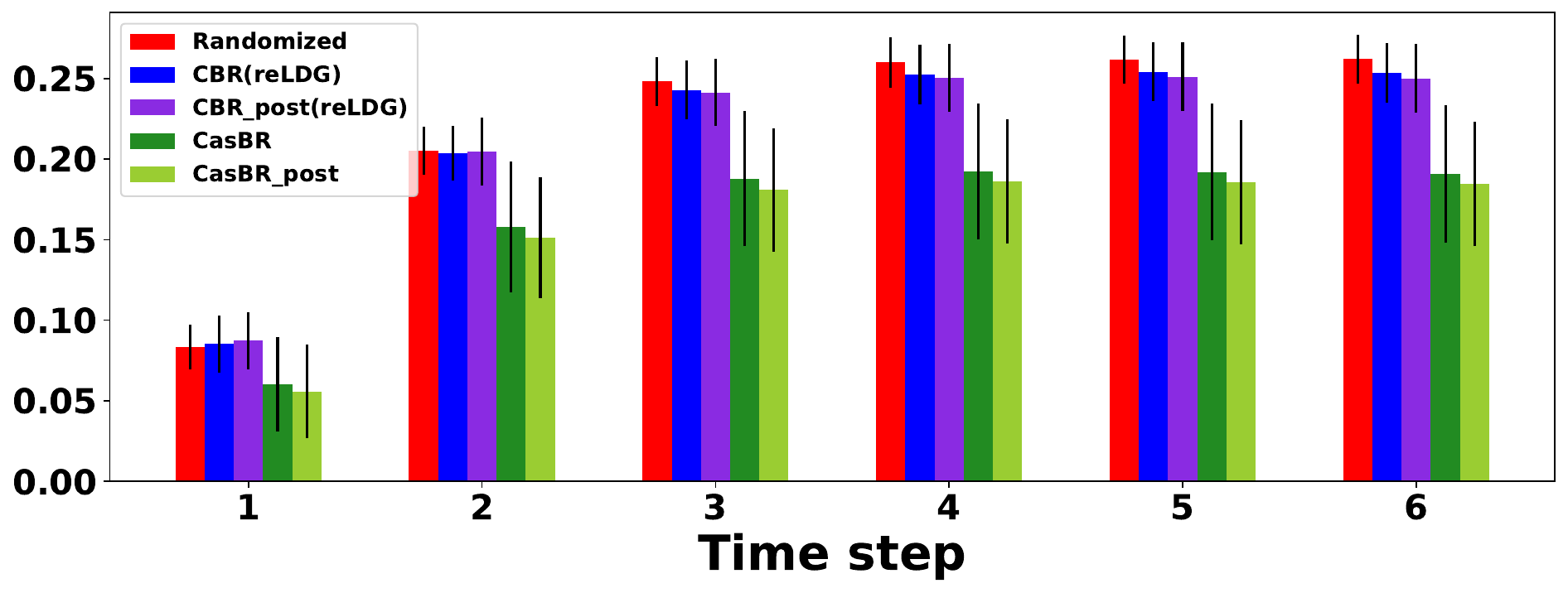}  
  \caption{Facebook}
  \label{fig:facebook}
\end{subfigure}
\caption{RMSE of total treatment effect estimation of different methods in consecutive time steps of the cascade propagation; Bars show the standard deviation of the estimated effects. Seed nodes are selected using the random sampling method. CasBR and CasBR-post get the lowest estimation error in different time steps.}
\label{fig:cascade}
\end{figure*}

\subsection{Baselines}
We compare the causal effect estimation error of five different approaches in our experiments:
\begin{itemize}
    \item \textbf{Randomized}: In this approach, nodes are randomly assigned to treatment or control independently. 
    \item \textbf{CBR}: In the Cluster-based Randomization (CBR) approach, a graph clustering algorithm is deployed to partition the graph nodes into densely-connected clusters, and treatment assignment at the cluster level dictates the node assignment within each cluster \citep{saveski-kdd17}.
   We leverage two different graph clustering algorithms: 1) \textit{Restreaming Linear Deterministic Greedy (reLDG)}, which generates a balanced proportion of nodes across partitions  \cite{nishimura-kdd13}, and 2) \textit{METIS}, which is a widely-used batch graph clustering algorithm that generates clusters with a minimal number of cross-partition edges ~\citep{karypis-jpdc98}. Both algorithms allow the experimenter to specify the number of clusters. reLDG has been used in prior studies to mitigate interference in network experiments \cite{saveski-kdd17,pouget-biometrika19}. In our experiments, CBR(reLDG) and CBR(METIS) represent CBR approaches utilizing reLDG and METIS clustering algorithms, respectively.
    \item \textbf{CBR-post}: This approach is a variant of the CBR method, but some nodes are excluded from the experiment using the post-processing technique described in Algorithm \ref{alg:icbrpost}. In our experiments, CBR-post(reLDG) and CBR-post(METIS) refer to CBR-post approaches using reLDG and METIS clustering algorithms, respectively.
    \item \textbf{CasBR}: This method is our proposed approach described in Algorithm \ref{alg:icbr}.
    \item \textbf{CasBR-post}: This is a variant of the CasBR approach where the post-processing technique described in Algorithm \ref{alg:icbrpost} is applied to the output of the CasBR method.
\end{itemize}
In CBR and CBR-post methods with both METIS and reLDG algorithms, we set the number of clusters equal to the number of cascade seed nodes in each experiment. 

We chose the alphas that led to the lowest RMSE, $\alpha=0.01$ in all datasets except Facebook where $\alpha=0.1$ 
The percentage of nodes excluded from the experiments during post-processing of each dataset using CasBR and CBR(reLDG) is shown in Table \ref{tb:post_pro}. We observe that CBR(reLDG) has a higher percentage of nodes with a more similar number of treatment and control neighbors than CasBR, indicating a higher level of interference in the CBR(reLDG) approach.
It is worth noting that our causal effect estimator only takes into account the treatment and control nodes. However, we do allow for the peer effects of bystander nodes on treatment and control nodes during TTE estimation. We set the unallowable peer effects from each bystander node to treatment or control nodes to 0.02.

\begin{figure*}[ht]
\centering
  \includegraphics[width=\linewidth]{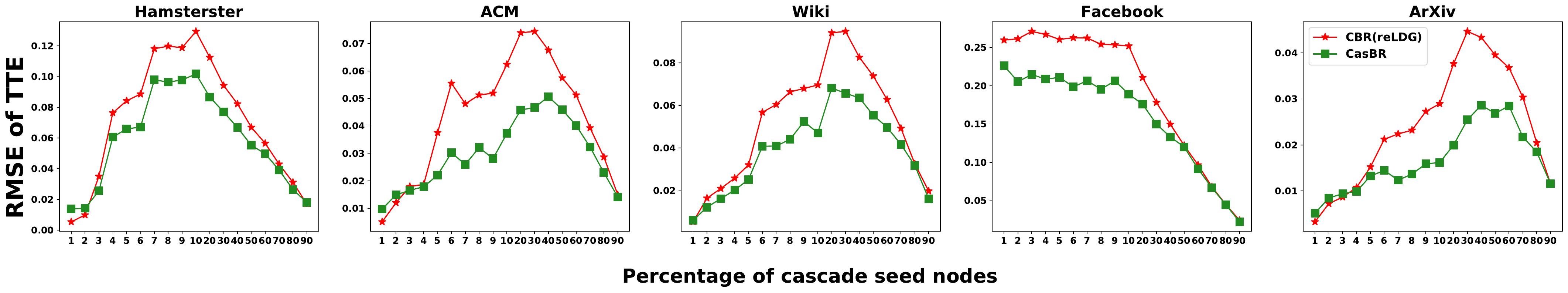}
\caption{ Comparison between RMSE of total treatment effect estimation of CasBR and CBR(reLDG); cascade seed nodes are selected using the random sampling approach. The number of clusters in CBR(reLDG) is equal to the number of cascade seeds.
}
\label{fig:random}
\end{figure*}
\begin{figure*}[ht]
\centering
  \includegraphics[width=\linewidth]{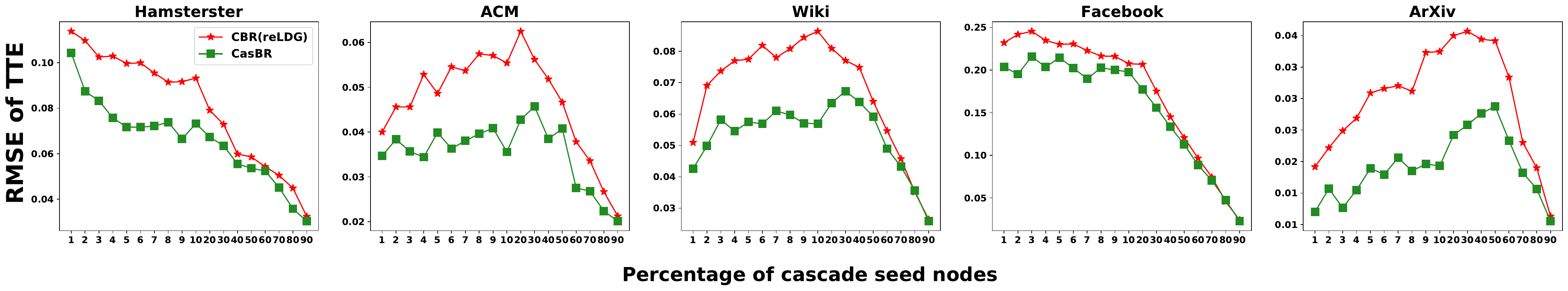}
\caption{ Comparison between RMSE of total treatment effect estimation of CasBR and CBR(reLDG); cascade seed nodes are selected using the NewGreedyIC approach. The number of clusters in CBR(reLDG) is equal to the number of cascade seeds.
}
\label{fig:greedy}
\end{figure*}

\subsection{Results}
\subsubsection{Interference Evaluation.} We evaluate the performance of different methods in mitigating interference. Fig. \ref{fig:interference} presents the comparison between RMSE of total treatment effects estimated by different frameworks in the last time step of the cascade propagation in all datasets. In this experiment, we use random sampling
 to select $10\%$ of the nodes as the cascade seed nodes. As expected, the CBR(reLDG) and CBR(METIS) methods outperform the Randomized approach in all datasets, except in the ACM dataset. The results suggest that the estimation error for CBR(reLDG) and CBR(METIS) is similar, except in the case of the Barab$\acute{a}$si-Albert network, where CBR(reLDG) outperforms CBR(METIS).
Furthermore, we observed that CasBR and CasBR-post produced significantly lower estimation errors compared to the baseline methods in all datasets, especially in Pubmed, Hateful Users, Facebook, Arxiv, Wiki, and Barab$\acute{a}$si-Albert networks, with estimated error reductions of $63.7\%$, $14.7\%$, $21.5\%$, $39.8\%$, $35.2\%$, and $33\%$, respectively, compared to the CBR(reLDG) approach. These findings remained consistent when varying the probabilities of spillover between treatment and control, as depicted in Fig. \ref{fig:interference_prob}.

\begin{figure}[t]
\centering
  \includegraphics[width=\linewidth]{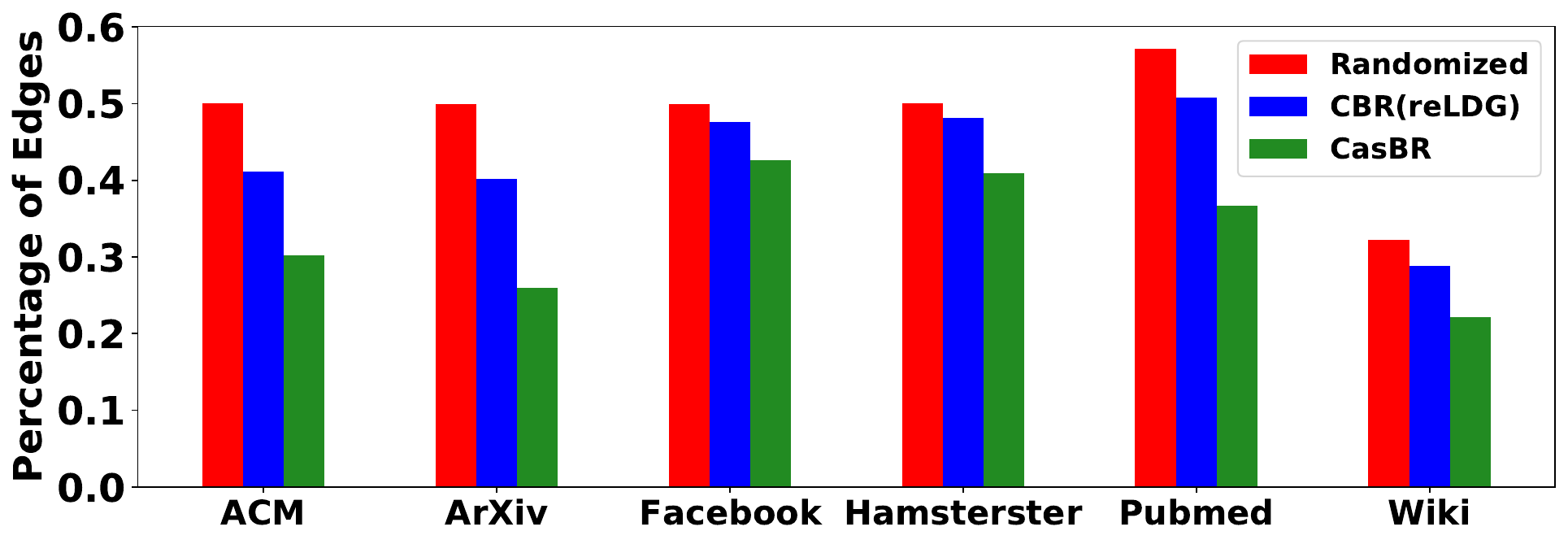}
\caption{ Percentage of edges between treatment and control nodes.
In all datasets, $10\%$ of the nodes are considered as the cascade seeds.
}
\label{fig:TCEdges}
\end{figure}

We also evaluate the percentage of edges connecting treatment and control nodes as an indicator of interference between these two groups. In this experiment, a random sampling technique is employed to choose 10\% of graph nodes as the seeds for the cascade. As illustrated in Fig. \ref{fig:TCEdges}, CasBR consistently demonstrates the lowest number of edges between treatment and control nodes across all datasets. In contrast, both the Randomized and CBR(reLDG) methods consistently exhibit the highest number of such edges.

Results show that applying the post-processing algorithm on CasBR, CBR(reLDG), and CBR(METIS) can improve the causal effect estimation error in all datasets. For example, the post-processing method reduces the estimation error of CasBR by $8\%$ in Hamsterster, $11.8\%$ in Arxiv, $9.3\%$ in Wiki, and $8.3\%$ in Barab$\acute{a}$si-Albert networks, and the estimation error of CBR(reLDG) by 
$7.6\%$ in PubMed and $2.2\%$ in Facebook dataset.
Given the similar performance of CBR(reLDG) and CBR(METIS), we use CBR(reLDG) exclusively in the subsequent experiments.

\subsubsection{Interference Evaluation Over Time.}
In this experiment, we assess the estimation error of CasBR and CBR(reLDG) in different time steps of cascade propagation. 
As represented in Fig \ref{fig:cascade}, the interference increases significantly over the cascade growth, with Randomized yielding an increase from $0.06$ to $0.24$, CBR(reLDG) from $0.05$ to $0.24$, and CasBR from $0.03$ to $0.2$ in the Hateful Users dataset, and Randomized yielding an increase from $0.08$ to $0.26$, CBR(reLDG) from $0.08$ to $0.24$, and CasBR from $0.06$ to $0.19$ in the Facebook dataset. Nonetheless, our experiments on both datasets show that CasBR and CasBR-post consistently outperform other baselines in all time steps of the cascade propagation. For instance, in the first time step of the cascade in the Hateful Users dataset, CasBR reduces the estimation error by $42.9\%$ compared to CBR(reLDG), and this reduction increases to $14.8\%$ in the last time step. The findings hold true for the Facebook dataset.

\subsubsection{Sensitivity to the Number of Cascade Seed Nodes.}
In this experiment, we explore the estimation error of CasBR vs. CBR(reLDG) when varying the number of cascade seed nodes. Considering the random sampling approach, Fig. \ref{fig:random} represents that CasBR gets a better performance compared to CBR(reLDG) in all datasets.
In addition, by increasing the number of cascade seeds which represents the number of clusters in the CBR(reLDG) method, the difference between the estimation error of CasBR and CBR(reLDG) increases, especially in ACM and ArXiv datasets.
Moreover, by increasing the number of cascade seeds, the estimation error for TTE in both methods follows a pattern of initial increase followed by a subsequent decrease, except in the Facebook dataset where it shows a decrease in the estimation error. For a high number of cascade seed nodes, the network quickly saturates, leading to the eventual convergence of error rates for both methods.

In contrast to employing random sampling, we employ the NewGreedyIC algorithm for the purpose of selecting the initial nodes for the cascade. Our results, presented in Fig. \ref{fig:greedy}, align with the previous findings that CasBR outperforms CBR(reLDG). However, in the Hamsterster dataset, we observe a decreasing trend in TTE estimation error when increasing the number of cascade seed nodes. These results highlight the impact of network properties on the amount of interference between treatment and control nodes, particularly in the context of selecting influential nodes. The set of influential nodes selected by the NewGreedyIC algorithm varies for different percentages, and increasing the percentage may exclude certain nodes from the set. Therefore, a model's performance may be influenced by the network structure and ego network of the influential nodes.

\section{Discussion}
Cascades of influence are natural phenomena frequently encountered in social systems experimentation. While previous studies have often advocated for clustering as the preferred method to mitigate interference in network experiments, this paper questions this approach by demonstrating that clustering may not be the best strategy when dealing with cascades of influence. Instead, we propose a network experiment design, CasBR, that leverages knowledge about cascade seed nodes to propagate treatment assignments to their multi-hop neighbors, thus ensuring that both the seeds and a significant portion of their multi-hop neighbors belong to the same treatment group.

Our empirical results demonstrate the better performance of CasBR compared to the baselines.
While in some datasets, e.g., PubMed and ArXiv, the RMSE seems to be small, it is important to also consider the effect size, because even a small error in the estimation can result in suboptimal decision-making. In large-scale marketing experiments, small effects (e.g., $<1\%$) can have significant implications for decision-making \cite{blake-ec14,fradkin-ms23}. A low RMSE (e.g., 0.01) indicates precise predictions, facilitating the detection and estimation of small effects. However, if the RMSE increases (e.g., 0.03), model accuracy decreases, and small effects may be suppressed by noise, leading to potential misinterpretation and missed opportunities for improvement.

Additionally, our findings highlight the impact of the number of active cascade seed nodes on the accuracy of causal effect estimation across various methods. We observe that by increasing the number of cascade seeds, the causal effect estimation error in both methods follows a pattern of initial increase followed by a subsequent decrease. However, as the number of activated cascade seed nodes increases substantially, the network experiences rapid saturation. This saturation ultimately leads to the convergence of performance between CBR and CasBR methods.

It is important to acknowledge that our proposed method relies on the availability of information about cascade seed nodes.
This is true, for example, when we know the early adopters of a product (e.g., people who bought the newest iPhones).
However, when such information is not unknown and cannot be easily inferred, our approach would not be applicable. 

One natural extension of this work is the development of network experiment frameworks for other models of information diffusion, such as the linear threshold model. 
Another possible extension is designing an experiment for scenarios in which the likelihood of neighbor activations can be inferred from cascades that already have been observed in the network. Addressing selection bias, i.e., the problem that treatment and control nodes can represent different populations of individuals~\cite{fatemi-icwsm20}, in network experiments with cascades is also a fruitful future direction.

\section{Acknowledgments}

% \fixme{RA to EZ: confirm the grant title/number}

This material is based on research sponsored in part by NSF under grant No.\@ 2047899, and DARPA under contract number HR001121C0168.

\bibliography{aaai23}
\end{document}